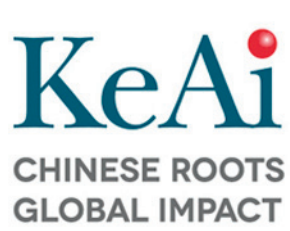

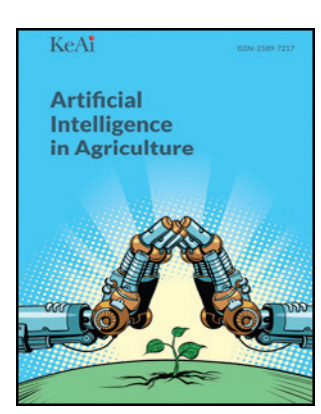

# Comparing YOLOv8 and Mask R-CNN for instance segmentation in complex orchard environments

Ranjan Sapkota *, Dawood Ahmed, Manoj Karkee *

*Center for Precision & Automated Agricultural Systems, Washington State University, 24106 N Bunn Rd, Prosser, 99350 Washington, USA*

ARTICLE INFO

ABSTRACT

Instance segmentation, an important image processing operation for automation in agriculture, is used to precisely delineate individual objects of interest within images, which provides foundational information for various automated or robotic tasks such as selective harvesting and precision pruning. This study compares the one-stage YOLOv8 and the two-stage Mask R-CNN machine learning models for instance segmentation under varying orchard conditions across two datasets. Dataset 1, collected in dormant season, includes images of dormant apple trees, which were used to train multi-object segmentation models delineating tree branches and trunks. Dataset 2, collected in the early growing season, includes images of apple tree canopies with green foliage and immature (green) apples (also called fruitlet), which were used to train single-object segmentation models delineating only immature green apples. The results showed that YOLOv8 performed better than Mask R-CNN, achieving good precision and near-perfect recall across both datasets at a confidence threshold of 0.5. Specifically, for Dataset 1, YOLOv8 achieved a precision of 0.90 and a recall of 0.95 for all classes. In comparison, Mask R-CNN demonstrated a precision of 0.81 and a recall of 0.81 for the same dataset. With Dataset 2, YOLOv8 achieved a precision of 0.93 and a recall of 0.97. Mask R-CNN, in this single-class scenario, achieved a precision of 0.85 and a recall of 0.88. Additionally, the inference times for YOLOv8 were 10.9 ms for multi-class segmentation (Dataset 1) and 7.8 ms for single-class segmentation (Dataset 2), compared to 15.6 ms and 12.8 ms achieved by Mask R-CNN's, respectively. These findings show YOLOv8's superior accuracy and efficiency in machine learning applications compared to two-stage models, specifically Mask-R-CNN, which suggests its suitability in developing smart and automated orchard operations, particularly when real-time applications are necessary in such cases as robotic harvesting and robotic immature green fruit thinning.

## 1. Introduction

Instance segmentation is a powerful computer vision technique that combines the benefits of both object detection and semantic segmentation (Hafiz and Bhat, 2020). One of the key benefits of instance segmentation in agricultural applications is its ability to accurately quantify plant and crop structures (Zhang et al., 2020a, 2020b), which can provide valuable information about plant growth, disease identification, and yield estimation, and can provide a foundation for various key areas of research and development such as robotic green (immature) fruit thinning (Champ et al., 2020). Instance segmentation can provide precise measurements of plant features, such as leaf area, stem length, and plant height, with a high level of accuracy and efficiency (Chen et al., 2019; Lüling et al., 2021). The traditional methods of instance segmentation in agricultural images were mostly based on handcrafted features and classical image processing techniques such as Watershed Transform (Niu et al., 2016), Graph-based Segmentation (Pham and Lee, 2015), Active Contours (or Snakes) (Clement et al., 2013), level set (Gao et al., 2011; Jothiaruna et al., 2019; Ma et al., 2017), Region Growing (Gao et al., 2011; Jothiaruna et al., 2019; Ma et al., 2017), Morphological Operation (Gupta et al., 2017; Khirade et al., 2015) and Clustering-based methods (Arsan and Hameez, 2019; Tian et al., 2019a, 2019b). However, these methods require a lot of manual setups and refinements, making them time-consuming and less reliable (Ngugi et al., 2021). Additionally, these methods could not easily learn from new data, making them less flexible and difficult to adapt to different scenarios. Moreover, these methods involved multiple disjointed image processing stages such as noise removal, contrast adjustment, image enhancement, refinement and manually defining and/or extracting specific features such as edge, texture, or colors.

Transitioning from traditional image processing methods to deep learning-based techniques in instance segmentation represents a significant evolution in the analysis of agricultural imagery. Traditional methods such as Watershed Transform (Zeng et al., 2009), Graph-based Segmentation, and Active Contours rely heavily on predefined algorithms that segment images based on intensity gradients, colour, texture or connectivity, which often necessitates extensive manual tuning to extract the specific characteristics of different crops or conditions (Jayanthi and Shashikumar, 2019; Zeng et al., 2009). These techniques, while foundational in the early days of computer vision, require iterative adjustments and are limited by their inability to dynamically learn from new data or adapt to varied environments. In contrast, deep learning models, especially convolutional neural networks (CNNs), introduce layers of learning filters that automatically extract and learn the most informative features from vast amounts of data. Unlike traditional methods that depend on manual feature specification and are thus prone to human bias and error, deep learning systems learn to identify underlying patterns and irregularities in data/images, making them more robust and accurate. This capability is particularly beneficial in agricultural applications (e.g., canopy instance segmentation) where the variability in plant appearance and environmental conditions can be high (Coulibaly et al., 2022). CNN-based models offer end-to-end learning techniques, which not only reduces the processing time but also enhances the adaptability of the models to new, unseen scenarios, an essential feature for generalizable and scalable agricultural applications. This dynamic learning ability is a significant improvement over traditional methods that are static and constrained by their algorithmic rigidity.

* Corresponding authors.
*E-mail addresses:* ranjan.sapkota@wsu.edu (R. Sapkota), manoj.karkee@wsu.edu (M. Karkee).

More specifically, DL network architectures, including U-Net (Siddique et al., 2021), Mask R-CNN, and YOLO (Redmon et al., 2016), are increasingly utilized for a range of applications in agriculture. A key advantage of these DL techniques is their end-to-end learning approach, which enables direct mapping of raw images to segmentation results, thus enhancing consistency and reliability. Furthermore, transfer learning techniques allow for the adaptation of models pre-trained on extensive datasets to specific agricultural tasks, reducing both training times and data requirements. Utilizing these features of DL models, various agricultural applications have been investigated including plant disease identification (Rashid et al., 2023), yield prediction (Maji et al., 2022), pest detection (Lippi et al., 2021; Liu and Wang, 2020), soil health assessment (Qu et al., 2023), crop maturity analysis, and site-specific weed control application (Hu et al., 2021; Li et al., 2022; Wang et al., 2022) showcasing their versatility and efficiency in modern agricultural practices.

As mentioned before, instance segmentation techniques have been widely applied to crop disease management (Chen et al., 2021a, 2021b). Early detection of plant diseases is crucial for maintaining crop yield and quality. Utilizing instance segmentation, researchers can quantify symptoms such as leaf spots and discoloration and monitor the progression of these diseases over time (Tian et al., 2020). This capability is instrumental in developing effective disease management strategies, including targeted treatments and breeding for disease-resistant cultivars. Instance segmentation has also been proved pivotal for precise crop yield estimation. Accurate yield estimation is essential for growers and breeders to make informed decisions about crop management and to select traits for breeding new cultivars. Instance segmentation techniques can be used to accurately counting and sizing individual fruits or other canopy objects from images. Such information facilitates a precise yield estimation and provides key insights into cultivar characteristics (Chen et al., 2021a, 2021b). Past studies have demonstrated the effectiveness of these techniques in various relevant applications such as the segmentation of apple flowers (Tian et al., 2020), segmentation and localization of strawberry fruits for harvesting segmentation and counting of cranberries, and the segmentation of guava fruits and branches (Lin et al., 2021). The data derived from these studies assist in optimizing crop management strategies, including optimal application of water and fertilizers, and identifying high-yielding cultivars (Tian et al., 2020) as mentioned before.

In addition, instance segmentation has been applied extensively to develop machine vision systems for agricultural robots because it provides capabilities for robots to detect, delineate, and track individual objects of interest in agricultural fields using images or videos, such as fruits, branches, flowers, vegetables, and livestock (Jha et al., 2019). Detecting and tracking plant parameters such as leaves, stem, trunk, branch, flower, and fruit is necessary for a robot to automatically perform various tasks such as harvesting, and canopy, and crop-load management operations. In the last few years, several studies have implemented the use of deep learning-based instance segmentation techniques for developing robotic solutions for various agricultural applications such as tree pruning in dormant season picking fruits and vegetables (Jia et al., 2020; Yu et al., 2019; Zu et al., 2021), thinning flowers and fruitlet and identifying and killing weeds (Xie et al., 2021) among others.

Among the broad applications of deep learning techniques in agriculture, there has been a focus on the use of two specific architectures: YOLO (You Only Look Once) and Mask Region-Based Convolutional Neural Network (Mask R-CNN). These models, known for their effectiveness in instance segmentation, have been pivotal in advancing tasks such as crop detection, pest and disease management, weed identification, tree canopy segmentation, and canopy object (e.g., branch and fruit) detection. These tasks, critical in precision and automated agriculture, benefit immensely from the capabilities of these two deep learning models. As mentioned before, many recent studies conducted in agricultural applications used Mask-R-CNN-based (He et al., 2017) instance segmentation for tasks such as crop detection (Ganesh et al., 2019; Wang et al., 2021), pest and disease detection (Afzaal et al., 2021; Lin et al., 2020; Rehman et al., 2021), weed detection (Osorio et al., 2020), tree canopy segmentation (Safonova et al., 2021; Zhao et al., 2018), and tree branch detection (Safonova et al., 2021; Zhao et al., 2018). Concurrently, the YOLO family of models has been used widely in object detection because of its ability to handle tasks like object detection, image classification, and instance segmentation simultaneously with one-stage networks. Unlike Mask R-CNN, a two-stage model suitable for segmentation tasks (Soviany and Ionescu, 2018), YOLO optimizes the overall processing ensuring speed and efficiency crucial for real-time applications in agriculture such as robotic pruning, thinning (Hussain et al., 2023), and pesticide application (Seol et al., 2022).

A number of recent studies focused on the segmentation of tree trunks and branches, employing various deep learning approaches. For example, (Fu et al., 2023; Ma et al., 2021) used deep learning for automatic branch detection in jujube trees and (Zhang et al., 2018) used Regions-Convolutional Neural Networks (R-CNN) models alongside depth features for branch detection in fruiting wall apple trees. Segmenting plant canopy parts in dormant grapevines has also been studied widely using different deep learning techniques (e.g., (Guadagna et al., 2023)). Other models like ViNet (Gentilhomme et al., 2023) have emerged, providing deep learning solutions for estimating grapevine structures. Further advancements include the application of deep learning and geometric constraints for obscured branch segmentation and three-dimensional reconstruction (Kok et al., 2023), as well as the use of space colonization algorithms for dormant pruning in jujubee plants (Fu et al., 2023). A deep learning-based sensing system (called SPGnet) for jujubee plant by Baojian et al. (Ma et al., 2021), branch detection in apple trees using R-CNN by Zhang et al., 2018, and Lin et al.'s bab (Lin et al., 2021) tiny Mask R-CNN for guava branch reconstruction are other recent studies in this field. Additionally, Aguiar et al. (Aguiar et al., 2021) explored trunk segmentation using a semantic segmentation-based deep learning approach with a Single Shot Multibox Detector (SSD). In comparison with the performance measures reported by these latest, innovative methodologies available in the literature, YOLOv8 model presented in this study performed better in segmenting tree trunks in terms of both precision (0.95), recall (0.97) and mAP@0.5(0.74). Furthermore, while the Mask R-CNN model achieved relatively lower performance relative to YOLOv8, its performance was comparable or better with many recent studies on trunk and branch detection including (Gentilhomme et al., 2023; Guadagna et al., 2023; Kok et al., 2023; Xiang et al., 2022; Zhang et al., 2018).

Both models are extensively studied, as discussed above and as shown in Table 1, highlighting 23 publications in the last 3 years focusing specifically on analyzing images of modern apple tree canopies.

Building upon this background of widespread application of YOLOv8 and Mask R-CNN models, the primary goal of this study is to systematically compare and evaluate the performance of these two models (YOLOv8 and Mask R-CNN) for instance segmentation tasks in modern, commercial apple orchards. Through this comprehensive comparison, this research aims to provide insights into the suitability, efficiency, and potential challenges associated with implementing each model in agricultural automation applications. To achieve this goal, the following specific objectives will be pursued in this study:

1. To compare the performances of YOLOv8 and Mask R-CNN models in segmenting single-class objects, specifically green apples (fruitlets), in images collected from variable orchard environments in the early growing season; and
2. To evaluate the capabilities of these two models in segmenting multi-class objects, specifically primary branches and tree trunks of apple trees in images collected from a model apple orchard during the dormant season.

The comparison between YOLOv8 and Mask R-CNN in this study is founded on the significant advancements in the YOLO architecture that extend its capabilities beyond mere bounding box detection. Traditionally, while YOLO models were primarily known for their speed and efficiency in object detection, the latest iterations, particularly YOLOv8, have incorporated features supporting instance segmentation. This adaptation allows YOLOv8 to not only predict bounding boxes but also to generate precise object masks, aligning its functionalities more closely with those of Mask R-CNN, which has been a standard in instance segmentation. Therefore, comparing these two models is pertinent as both now offer robust solutions for instance segmentation, making the evaluation of their performance in agricultural applications, where both detection speed and segmentation accuracy are critical, highly relevant and scientifically justified.

The remainder of this paper is organized to provide a comprehensive comparison of YOLOv8 and Mask R-CNN models, specifically focusing on their application in commercial apple orchards. First, a “Background” section is presented to outline the theoretical frameworks of the one-stage (YOLOv8) and two-stage (Mask R-CNN) detectors, setting the stage for a deeper understanding of these complex models. Then a “Materials and Methods” section is provided to describe the experimental design, data acquisition, and analytical methodologies employed. Following the methodology, a “Results and Discussion” section is presented to report the results and critically discuss models' performance in various segmentation tasks, including their efficiency and efficacy. The paper concludes with a “Conclusion” section, which summarizes the research methods and the findings of the study. The paper ends with “Future Work” section, which summarizes the potential further comparison with state-of-the-art other models.

## 2. Deep learning models

Deep learning models used for object detection are generally categorized into two distinct approaches: one-stage and two-stage detectors (Carranza-García et al., 2020). Two-stage detectors, such as Mask R-CNN, first generate regions of interest (ROIs) in an initial stage, using a Region Proposal Network (RPN) (He et al., 2017). These regions are then classified and refined in the second stage to provide precise object localization and classification. This approach is known for its high accuracy due to the focused refinement of detected objects. On the other hand, one-stage detectors like YOLO streamline this process by directly predicting object classes and bounding boxes in a single pass through the network, sacrificing some accuracy for significant gains in speed. One-stage models do not separate the detection into distinct region proposals and refinement stages, allowing them to operate faster, making them well-suited for applications requiring real-time processing (Carranza-García et al., 2020). Both methodologies have continued to evolve in recent years to address the trade-offs between speed and accuracy.

It is true that numerous one-stage and two-stage detectors have been developed over the past decade, each tailored for specific performance criteria in terms of speed and accuracy. Some of the most widely used two-stage detectors include Fast R-CNN (which improves the efficiency of feature usage), Faster R-CNN (which integrates a region proposal network), and R-FCN and Cascade R-CNN (which enhance localization and classification accuracy through specialized networks). Similarly, most widely used one-stage detectors include SSD (Single Shot MultiBox Detector) and RetinaNet (which introduced focal loss to handle class imbalance), and family of YOLO model including YOLOv5 and YOLOv6 (which emphasize detection speed). Despite the availability of numerous one-stage and two-stage models, YOLOv8 and Mask R-CNN have been the most widely used models in agricultural applications with highly impactful results (Duong-Trung and Duong-Trung, 2024; Jabir et al., 2023; Mu et al., 2023; Xu et al., 2020; Yang et al., 2023; Yu et al., 2022; Yue et al., 2023).Comparative studies have clearly demonstrated their superior performance in detecting and segmenting agricultural objects under varied conditions, as outlined in Table 1 of this manuscript. These models balance the trade-offs between speed, accuracy, and robustness, making them especially suitable for the dynamic environments encountered in agricultural settings. Based on these concaving past studies, Mask R-CNN (for its precise segmentation capability), and YOLOv8 (for its exceptional speed), have thus been selected for this study.

**Table 1**
Highlighting the studies conducted in the last three years on YOLO and Mask R-CNN during different apple orchard environments.

| References | Year | DL model | Objectives |
|---|---|---|---|
| (Chen et al., 2021a, 2021b; Wu et al., 2021) | 2021 | YOLO-V4 | Apple detection in a complex scene |
| (Huang et al., 2021; Liu et al., 2021) | 2021 | Mask R-CNN | Deep learning-based apple detection |
| (Huang et al., 2021; Kuznetsova et al., 2021) | 2021 | YOLO-V3 | Green fruit detection (apples, mangoes) |
| (Kuznetsova et al., 2021; Wang et al., 2021) | 2021 | YOLO-V5 | Apple fruitlet detection for fruitlet thinning |
| (Tong et al., 2022) | 2022 | Mask R-CNN | Branch identification and junction points localization in apple trees; Trunk identification and segmentation |
| (Gao et al., 2022; Zhang et al., 2022) | 2022 | YOLO-V4 | Apple detection, counting, and tree trunk tracking in modern orchards |
| (Lu et al., 2022) | 2022 | YOLO-V4 | Immature/mature apple detection on dense-foliage tree architectures for early crop-load estimation |
| (Lv et al., 2022) | 2022 | YOLO-V5 | Identification method for the apple growth pattern in the orchard |
| (Su et al., 2022) | 2022 | YOLO-V5 | Tree trunk and obstacle detection in apple orchards |
| (Jia et al., 2022) | 2022 | Mask R-CNN | Ripe and green apple segmentation in orchards |
| (Cong et al., 2022) | 2022 | Mask R-CNN | Tree and tree crown segmentation in orchards |
| (Karthikeyan et al., 2023) | 2023 | YOLO-V3 | Apple fruit quality detection |
| (Ma et al., 2023) | 2023 | YOLO-V7 | Detection and counting of small target apples |
| (Hussain et al., 2023; Jia et al., 2022) | 2023 | Mask R-CNN | Green apple segmentation |

### 2.1. Mask R-CNN

Mask R-CNN is a deep learning model designed for object detection and instance segmentation, renowned for its accuracy and efficiency. Its strength lies in its ability to precisely identify and delineate each object in an image, making it highly effective for complex image analysis tasks. The model was developed by researchers at Facebook AI Research in 2017 and builds on top of the Faster R-CNN object detection model by adding a branch for predicting object masks in parallel with the existing branch for bounding box detection (He et al., 2017). The architecture of Mask R-CNN consists of three main components: a backbone network, a region proposal network (RPN), and two parallel branches for bounding box detection and mask prediction as shown in Fig. 1. The backbone network is typically a convolutional neural network (CNN) that extracts features from the input images and is shared by both branches. The RPN generates a set of region proposals that are likely to contain objects, based on the feature maps generated by the backbone network. The bounding box branch predicts the class label and bounding box coordinates for each region proposal, while the mask branch predicts a binary mask for each object instance within the bounding box.

However, the application of Mask R-CNN and other deep learning models in agriculture comes with several challenges. First, the performance of the model heavily relies on the quality and diversity of the training dataset. Agricultural environments are highly variable, with changes in lighting, weather conditions, and plant growth stages, all of which can affect the model's accuracy (Hoogenboom, 2000). Moreover, Mask R-CNN requires substantial computational resources for training and inference (Zhang et al., 2020a, 2020b), which can be a limitation in real-time applications on the farm where such resources are limited.

Ongoing studies are focusing on addressing these challenges and optimizing deep learning models such as Mask R-CNN for improved efficiency and robustness. These efforts include integrating more adaptive and scalable neural network architectures, improving data augmentation techniques to make the model more resilient to environmental variabilities, and developing lightweight versions of the models that maintain high accuracy while being more resource efficient. For instance, Mask R-CNN has been deployed to identify specific picking points on tea plants (Wang et al., 2023a, 2023b) to aid robotic harvesting of quality tea-leaves while minimizing damage to plants. Furthermore, this model has shown promising results in various horticultural applications such as assessing the ripeness of strawberries (Tang et al., 2023). By combining Mask R-CNN with region segmentation techniques, the system effectively distinguished between different ripeness stages, enabling growers to optimize the timing of strawberry picking for better market prices and reduced waste. In apple orchards, Mask R-CNN has been utilized for flower detection and identification of the king flower, which is critical for targeted pollination strategies (Amogi et al., 2023). This application is expected to help improve pollination efficiency leading to higher fruit yield and quality. Mask R-CNN has also been applied to monitor crop stress such as estimating fruit surface temperature using IoT sensors. This technology allows for real-time monitoring and management of fruit crops, thus helping orchard managers to mitigate the effects of heat stress and maintain fruit quality and yield.

### 2.2. YOLOv8

The YOLO family of object detection and instance segmentation models have evolved rapidly over the last several years, with each new iteration introducing improvements in accuracy and/or speed. YOLOv8 (Fig. 2), the latest one-stage model, was built on the foundations provided by previous YOLO models, such as YOLOv3 and YOLOv5. Compared to two-stage models, YOLOv8 directly predicts bounding boxes and class probabilities without the need for a separate region proposal network, streamlining the object detection process. One key innovation in YOLOv8 is the adoption of an anchor-free, center-based approach for object detection, which offers several advantages over traditional anchor-based methods such as YOLOv5, YOLOv6, and YOLOv7. YOLOv8 implements Pseudo Ensemble or Pseudo Supervision (PS), a method that involves training multiple models with distinct configurations on the same dataset to generate a more diverse set of predictions, improving the accuracy and robustness of the final prediction. Additionally, YOLOv8 leverages the Darknet-53 architecture, a 53-layer deep convolutional neural network optimized for feature extraction and object detection. One significant architectural change in YOLOv8 is the replacement of the C3 module with the C2F module. The C3 module, also known as the convolutional module, processes input data through a series of convolutional operations. The C2F module, an improved version of the C3 module, enhances accuracy and processing times compared to previous models. Furthermore, YOLOv8 substitutes the 6 × 6 Convolutional (Conv) layer with a 3 × 3 Conv layer in the model backbone, reducing the number of parameters and creating a more compact, computationally efficient network. YOLOv8 also employs a decoupled head, which separates the tasks of predicting object presence and classifying object types, thereby improving both accuracy and processing speed. This refinement positions YOLOv8 as an effective solution for both object detection and instance segmentation in computer vision.

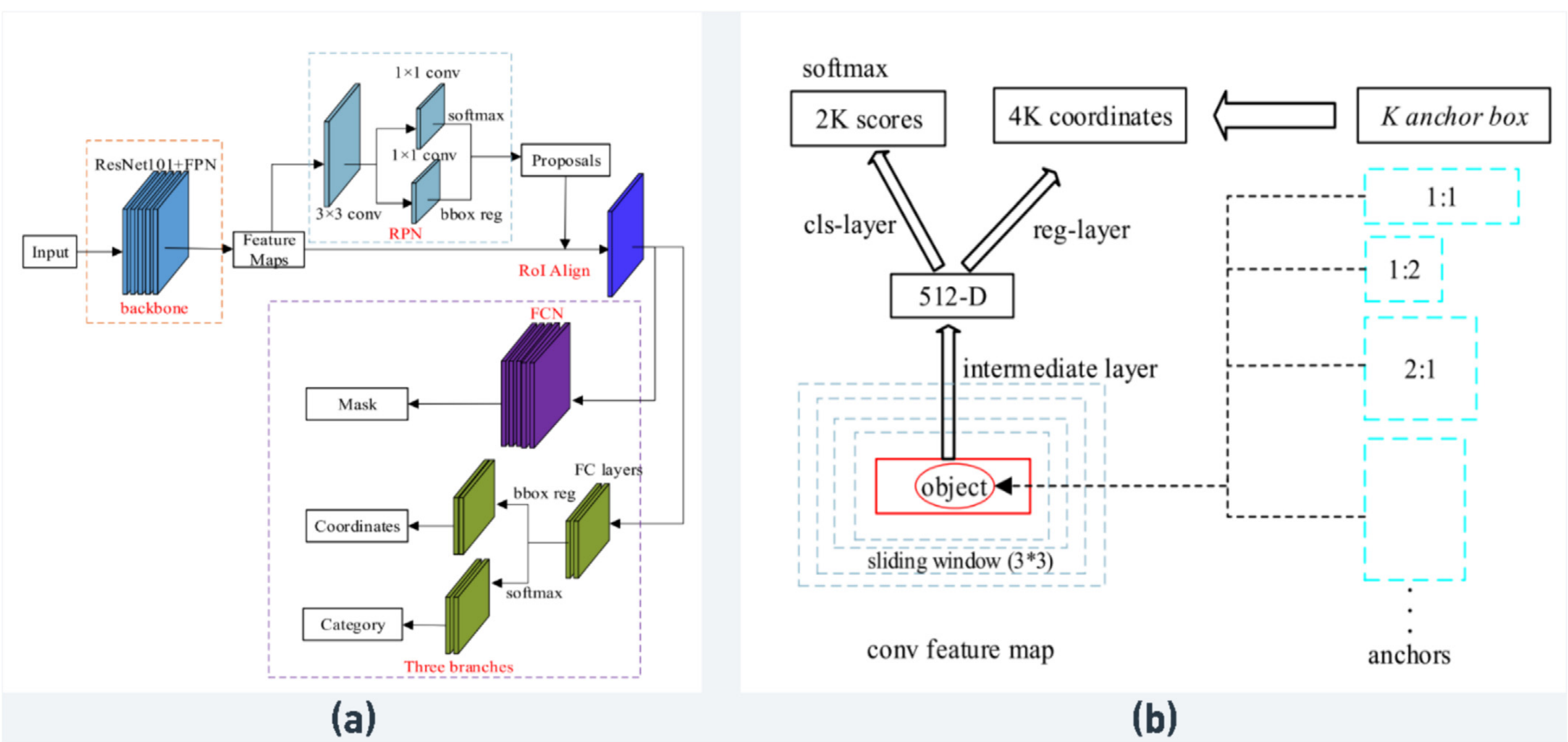

**Fig. 1.** Mask-R-CNN architecture with; (a) structure diagram, highlighting the backbone network, RPN, bounding box, and mask prediction branches; and (b) detailed view of the Region Proposal Network (RPN).

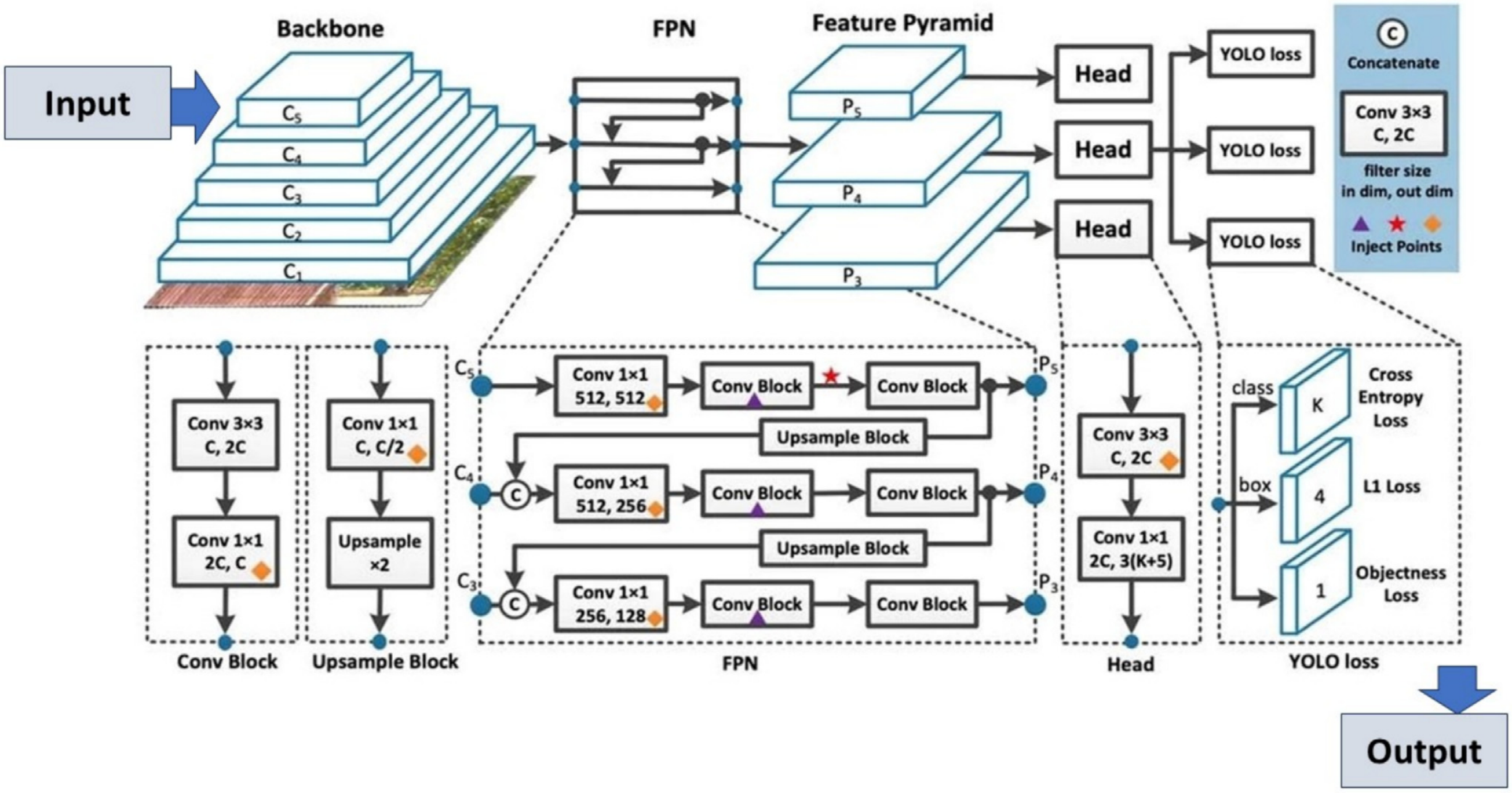

**Fig. 2.** YOLOv8 architecture showcasing its innovative design for object detection and segmentation (https://deci.ai/blog/history-yolo-object-detection-models-from-yolov1-yolov8/).

YOLOv8 offers several configurations to cater to different needs for computational speed and accuracy (Li et al., 2023; Wang et al., 2023a, 2023b): YOLOv8-Tiny for fast processing at the cost of some accuracy, ideal for real-time applications on limited-resource devices; YOLOv8-Small, which balances speed with more detailed detection capabilities; YOLOv8-Standard for robust performance in diverse settings; and YOLOv8-Large, which prioritizes high accuracy for critical applications where details and precision are paramount.

The recent advancements in YOLOv8 have facilitated its adoption in diverse agricultural applications, demonstrating effectiveness in addressing specific challenges inherent to various farming environments discussed earlier. By optimizing the processing of low-level features, YOLOv8 becomes a powerful tool for the early detection of subtle signs of agricultural pests and diseases, critical for maintaining crop health (Zhang et al., 2023). For example, enhanced versions of YOLOv8 have been applied to detect diseases in vegetables within greenhouse environments (Wang and Liu, 2024) ensuring early detection and management of plant diseases. Further innovations include the integration of attention mechanisms into YOLOv8 to enhance the object detection capabilities, which was tested by (Yang et al., 2023) to improve tomato detection accuracy in cluttered agricultural environments.

## 3. Materials and methods

This study consisted of four major steps as outlined in Fig. 3a beginning with RGB images acquisition from commercial orchards in two distinct seasons (Fig. 3b as dormant season and 3c as early growing season). These images, captured under varying environmental conditions such as bright and cloudy days, were then manually annotated to create the training and testing datasets. The training dataset was subsequently used to train the two deep learning models mentioned previously, and their performance in instance segmentation was evaluated using the test dataset.

### 3.1. Study site and data acquisition

This study was conducted in a commercial apple orchard (Fig. 3b and c) owned and operated by Allan Brothers Fruit Company, located at Prosser, Washington State, USA. The orchard was planted in 2009 with a Scilate apple cultivar with a row spacing of 9.0 ft., and a plant spacing of 3.0 f. and was trained to a V-trellis architecture. Two sets of RGB images were acquired using IntelRealsense 435i (Intel Corporation, California, USA); one in November 2022 creating dormant season dataset as shown in Fig. 3b and e, while the other set of images was acquired in June 2023 (just before manual fruitlet thinning) which provided the dataset for early growing season as illustrated in Fig. 3c and f. The Intel RealSense camera was selected for capturing RGB images due to its software development kit's (Intel RealSense SDK) ability to adjust parameters and capture high-quality images.

### 3.2. Data preparation

Two kinds of datasets comprising 1553 RGB images, capturing a variety in orchard lighting conditions were prepared for analysis of the deep learning model. Dataset 1 comprised 474 images from the dormant season, which were annotated manually to represent multi-class objects: the tree trunk and primary branches growing out from the trunks (Fig. 4). Altogether, 1141 annotations for the tree trunk and 2369 annotations for the tree branches were generated manually by creating the polygon over desired objects in these images using the image labeling software Labelbox. Likewise, dataset 2 comprised 1079 images from the green fruit growing season in which 5921 annotations of immature green apples were generated. During the image preprocessing stage using the label box software, all these annotations were formatted in accordance with the COCO dataset specification, which meets the requirement of both the YOLOv8 and Mask R-CNN model for image segmentation.

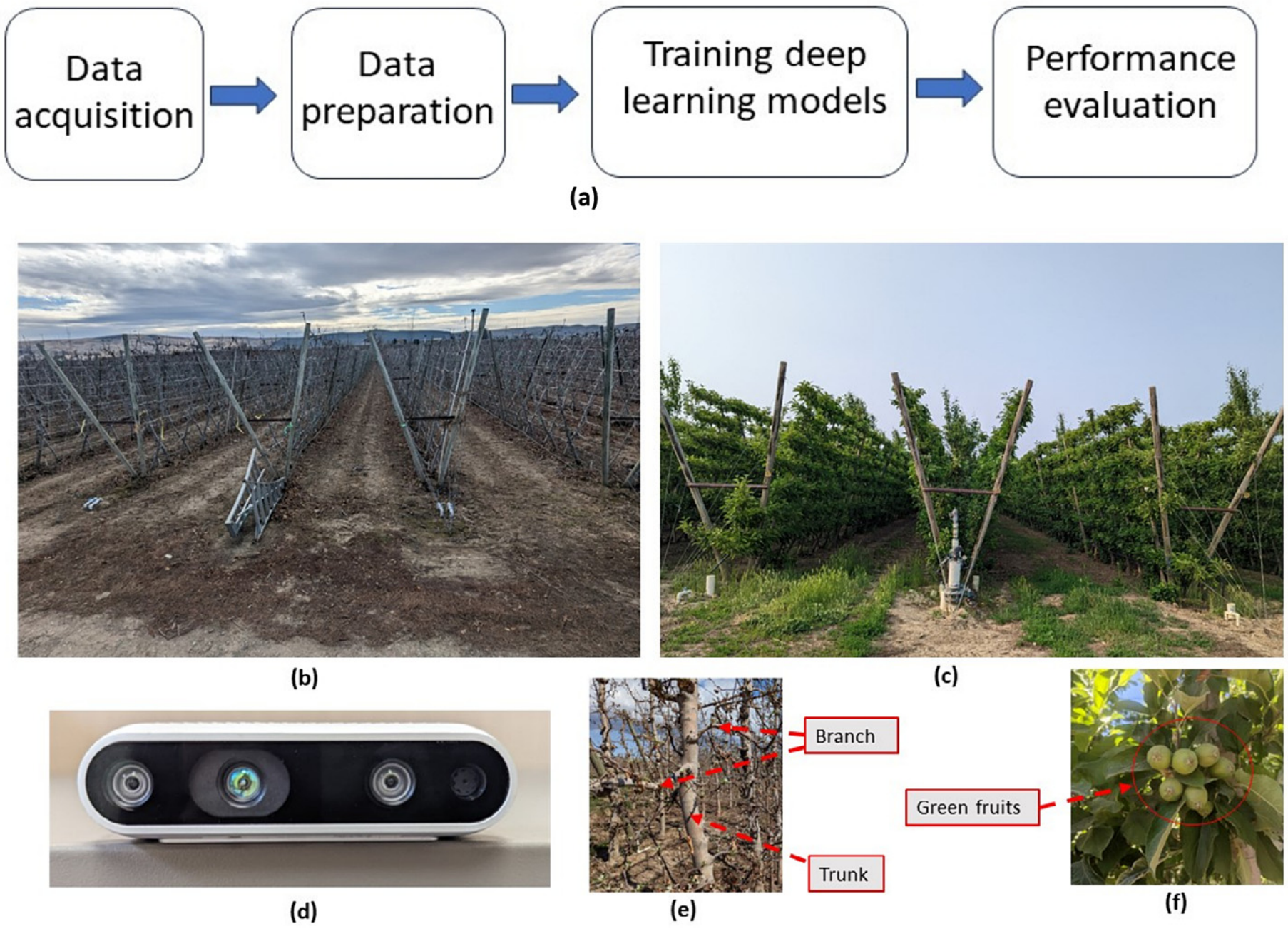

**Fig. 3.** (a) Overall workflow diagram used in this research; (b) An example image of an Apple orchard during the dormant season (November 22, 2022); (c) An example image of an apple orchard during early fruit growing season (June 18, 2023); (d) IntelRealsense 435i camera used to acquire images to train and test the instance segmentation models; (e) Example trunks and branches used to annotate the dormant season images; and (f) Example immature green fruits (fruitlets) used to annotate early growing season images.

Furthermore, to facilitate model training and validation, both datasets were resized to 640 × 640 pixels, and both datasets were divided randomly into training, validation, and test subsets, following an 8:1:1 distribution ratio for each object class.

### *3.3. Deep learning model implementation*

Both the YOLOv8 and Mask R-CNN models were trained on a workstation with an Intel Xeon® W-2155 CPU @ 3.30 GHz x20 processor, NVIDIA TITAN Xp Collector's Edition/PCIe/SSE2 graphics card, 31.1 GiB memory, and Ubuntu 16.04 LTS 64-bit operating system. The backend framework for the model implementation was Pytorch, operating on a Linux system. To optimize performance, the learning rate used was 0.001, the batch size used was 32, and the dropout rate used was 0.5 to mitigate overfitting. The training was conducted over 1000 iterations. The model training was stopping before reaching 1000 epochs, if the model performance did not improve for 20 consecutive epochs over the validation dataset, which was useful to minimize model overfitting to the training dataset and improving generality. An initial learning rate of 0.01 was used in training both models, whereas the momentum and weight decay used were 0.937 and 0.0005 respectively for the two models. These parameter settings were chosen to optimize the speed of the training process while minimizing the chances of overfitting the model to the training dataset. During the initial three epochs, a warm-up phase was employed, using a momentum of 0.8 and a bias learning rate of 0.1, to stabilize the model's optimization and mitigate the risk of being stuck at a poor local minimum.

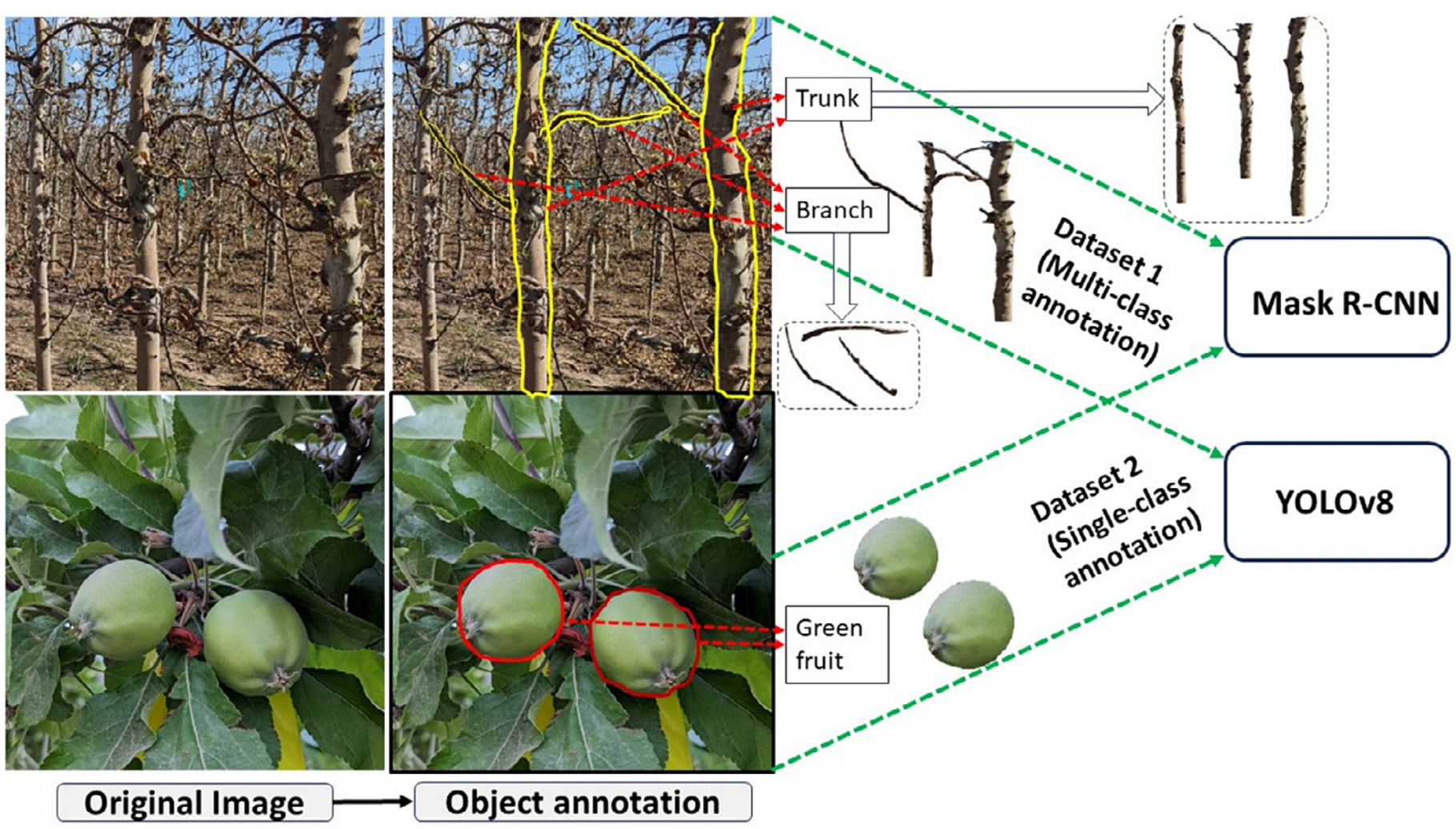

**Fig. 4.** Workflow diagram showing the two types of datasets used in the study; Dataset 1 included the dormant season apple trees with multi-class objects (Trunk and branch) and Dataset 2 included growing season apple tree canopies with immature green fruits.

During the training process, various augmentation techniques were applied to enhance model robustness and generalization such as hue augmentation (0.015), saturation augmentation (0.7), value augmentation (0.4), translation adjustments (0.1), scaling variations (0.5), and a 50% probability for left-right flips. Additionally, a mosaic augmentation was applied with a probability of 1.0. After the model training was completed, the model outputs were converted to TorchScript format to simplify further processing to evaluate the performances of both YOLOv8 and Mask R-CNN models in terms of precision, recall, mean average precision (mAP), and area under curve (AUC) as discussed below. The detailed the specific data augmentation techniques and hyperparameter values utilized during training is presented in Table 2:

The number of training epochs was determined through a preliminary test. In this test, the model performance with the validation dataset was monitored during the training process and the number of epochs when validation accuracy started to decrease or stay flat was chosen as the optimal number of epochs, which was expected to help avoid model overfitting.

### 3.4. Performance evaluation

To evaluate the instance segmentation capabilities of the Mask R-CNN and YOLOv8 models, five distinct metrics were used: Precision, Recall, mean Average Precision (AP) at 0.5 intersection over union (mAP@ 0.5 IOU), Area Under the receiver operating characteristic Curve (AUC), and Inference speed. Precision is defined as the proportion of correctly identified positive instances to the total predicted positive instances, as depicted by Eq. 1. Similarly, recall, depicted by Eq. 2, quantified the proportion of correctly identified positive instances out of all actual instances of the target objects. Furthermore, the mean average precision (mAP), represented as the average of the AP across k categories (Eq. 4), was crucial in evaluating the model's precision at a threshold of 50% overlap between predicted and true object boundaries/bounding boxes. The area under the curve (AUC), defined by Eq. 5, assessed the model's classification efficacy across all possible thresholds. The model's efficiency in processing and delivering predictions was measured by the inference speed and was inversely related to the time taken per image analysis.

These metrics are calculated as follows:

$$Precision = \frac{TP}{TP + FP} \tag{1}$$

**Table 2**
Data augmentation and regularization parameters used in training models in this study.

| Methods Applied | Value |
|---|---|
| Hue augmentation (fraction) | 0.015 |
| Saturation augmentation (fraction) | 0.7 |
| Value augmentation (fraction) | 0.4 |
| Rotation | 0.0 |
| Translation | 0.1 |
| Scale | 0.5 |
| Flip left-right (probability) | 0.5 |
| Mosaic (probability) | 1.0 |
| Weight decay | 0.0005 |

$$Recall = \frac{TP}{TP + FN} \tag{2}$$

$$IoU = \frac{Area\ Overlap}{Area\ Union} = \frac{TP}{FP + TP + FN} \tag{3}$$

$$mAP = \left(\frac{1}{K}\right) \sum_{i=0}^{k} (AP)i \tag{4}$$

$$\mathrm{AUC} = \int_0^1 TPR\left(FPR)^{-1}(u)\right) du \tag{5}$$

where TP, FP, and FN represent true positive, false positive, and false negative object instances respectively. Variable ‘k’ represents the total number of object classes, and $(AP)_i$ refers to the average precision calculated for the $i^{th}$ class among these k classes. AP is the area under the precision-recall curve for a given class. TPR represents the true positive rate, FPR is the false positive rate, and t indicates the time taken for the model to infer results for a given (single) image.

## 4. Results and discussion

### 4.1. Single-class object segmentation of immature green apples (fruitlets)

For single-class segmentation for immature green fruits, the Precision-Confidence curves depicted in Fig. 5, revealed that the YOLOv8 model achieved a maximum precision of 1.00 when the confidence threshold was 0.929 (Fig. 5a). Correspondingly, Recall-Confidence curves for the respective models are presented in Fig. 6, which showed that YOLOv8's recall reached 0.97 at the minimum confidence threshold of 0.000. This high recall rate, or sensitivity, indicates the model's ability to correctly identify a high percentage of actual objects, which showed models effectiveness in segmenting green fruits even at the lowest confidence levels. Additionally, YOLOv8 outperformed Mask R-CNN in terms of mean average precision (mAP), achieving 0.939 at a 0.5 IoU threshold for green fruits and overall categories, compared to mAP of 0.902 achieved with Mask R-CNN (Fig. 7).

The performance differences between YOLOv8 and Mask R-CNN generally reflected the distinct nature of their architectures and the way they process images. YOLOv8, being a one-stage detector, is designed for speed and accuracy, making it capable of excluding similar non-target areas, as observed in the segmentation tasks (Fig. 8b). Its direct approach to object detection avoids the region proposal step, leading to fewer false positives in areas of the canopy that resembled the target fruit in colour. Mask R-CNN, on the other hand, uses a two-stage process, which involved generating region proposals before classifying and segmenting objects. This can sometimes result in the inclusion of non-target areas, such as leaves and stems being misclassified as fruits (Fig. 8c). Moreover, its performance appears to be more sensitive to lighting variations, which can lead to errors in object identification under the extreme sides of lighting situation such as bright, direct sun-light and dark shadows (Fig. 9c).

Despite these differences, there are specific situations where Mask R-CNN could still be the preferred choice. Its two-stage process, particularly the region proposal step, can be advantageous in complex segmentation tasks where precision is critical, and objects are densely packed or partially obscured. In the past, green fruit segmentation has been investigated using various approaches. Wei et al.'s D2D framework (Wei et al., 2022), GHFormer's focus on night-time detection (Sun et al., 2022), Liu et al.'s FCOS model for obscured fruits (Liu et al., 2022), Jia et al.'s ResNet-based FoveaMask (Jia et al., 2021), and Sun et al.'s combination of GrabCut and Ncut algorithms (Sun et al., 2019) each offered solutions to specific segmentation challenges such as lower accuracy and higher computation cost. Some studies also explored semi-automated models (Prabhu and Rani, 2021). However, the performance of the YOLOv8 model in this study exceeded those of the reviewed past studies. Likewise, the performance of the Mask R-CNN model in segmenting immature green fruits, while not as high as YOLOv8's, still surpassed many recent approaches (Karthikeyan et al., 2023; Liu et al., 2022; Lu et al., 2022; Prabhu and Rani, 2021; Sun et al., 2022; Tian et al., 2019a, 2019b).

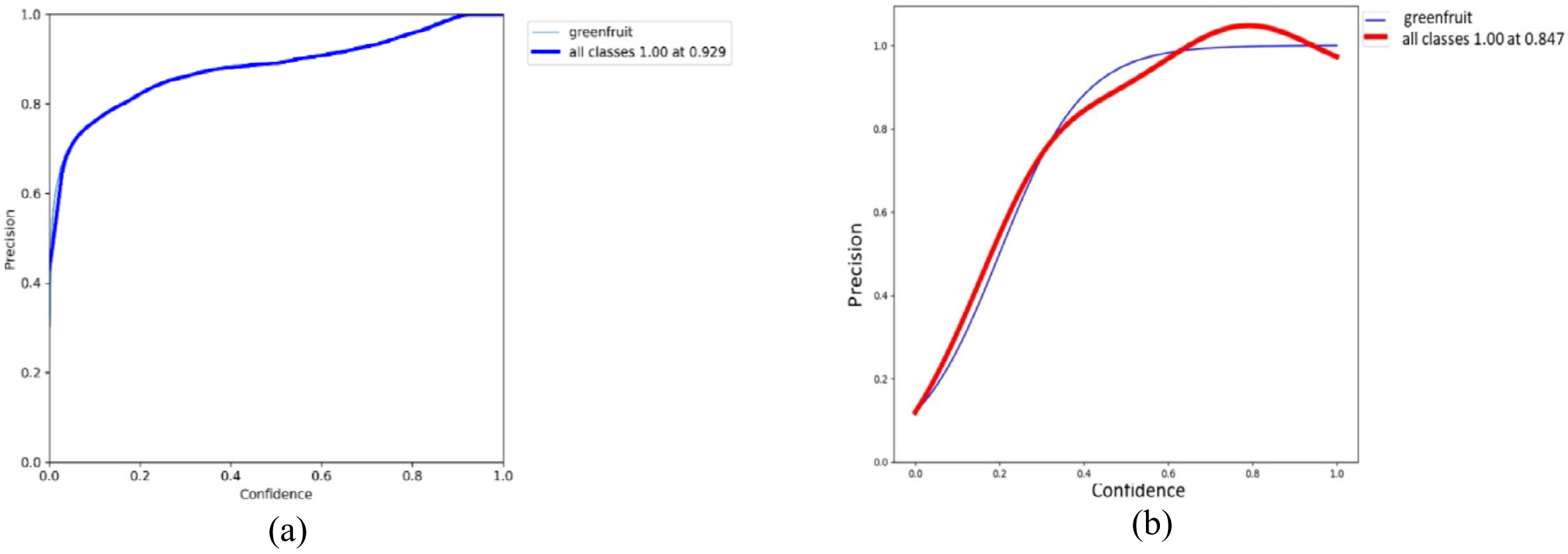

**Fig. 5.** Precision-Confidence curve for single class segmentation of immature green apples (fruitlets) using; (a) YOLOv8; and (b) Mask R-CNN.

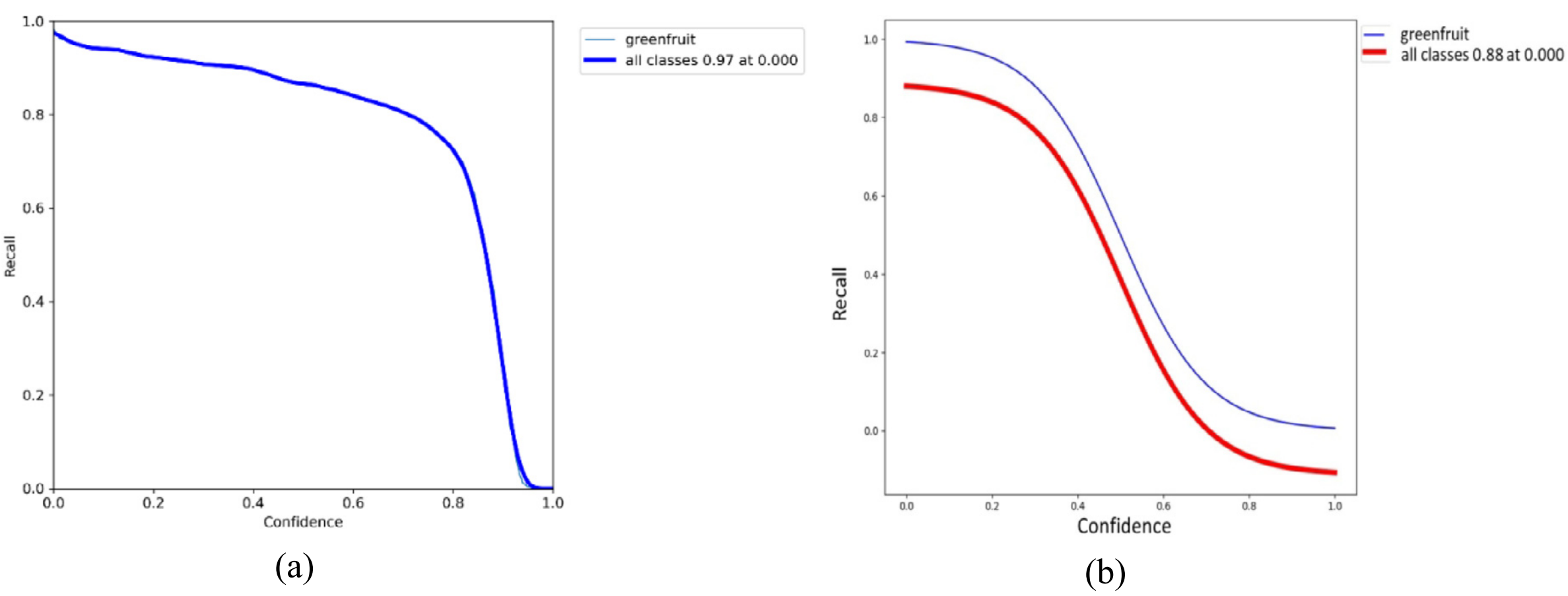

**Fig. 6.** Recall-Confidence curve for single class segmentation of green apple fruits using; (a) YOLOv8; and (b) Mask R-CNN.

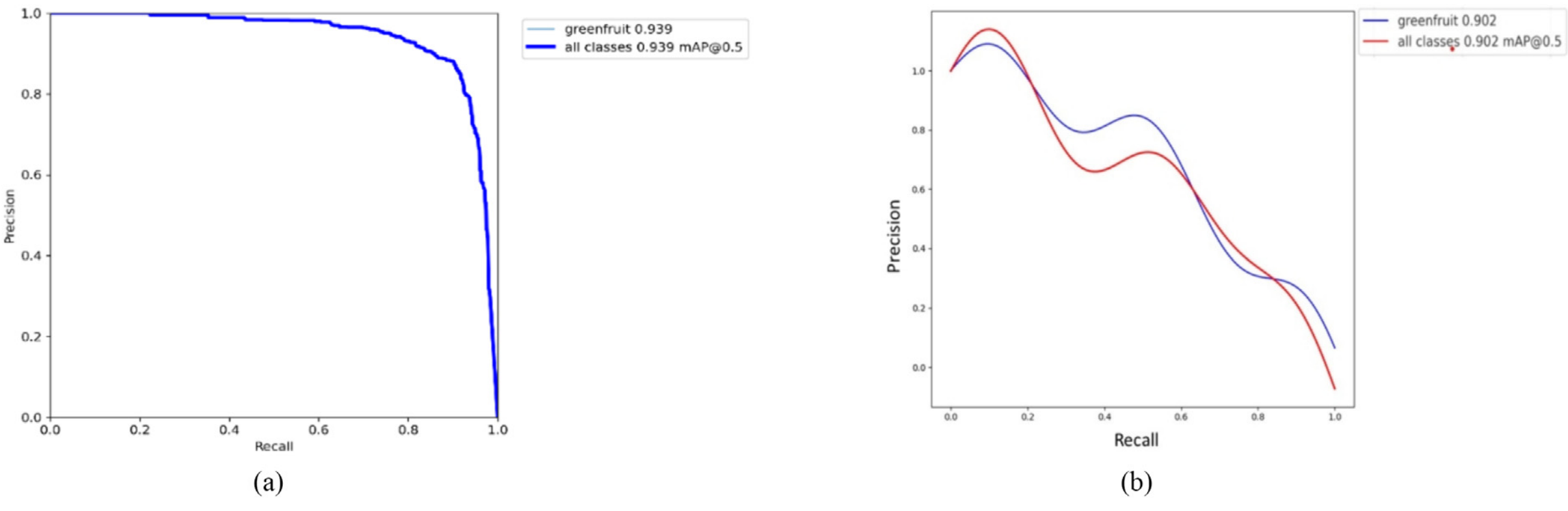

**Fig. 7.** Precision-Recall curve for single class segmentation of green apple fruits at mAP@0.5; (a) YOLOv8; and (b) Mask R-CNN.

Based on the recently published results such as multi-class fruit detection using a robotic vision system by Wan et al. (Wan and Sotirios, 2020), the authors compared YOLOv3, Faster-R-CNN and

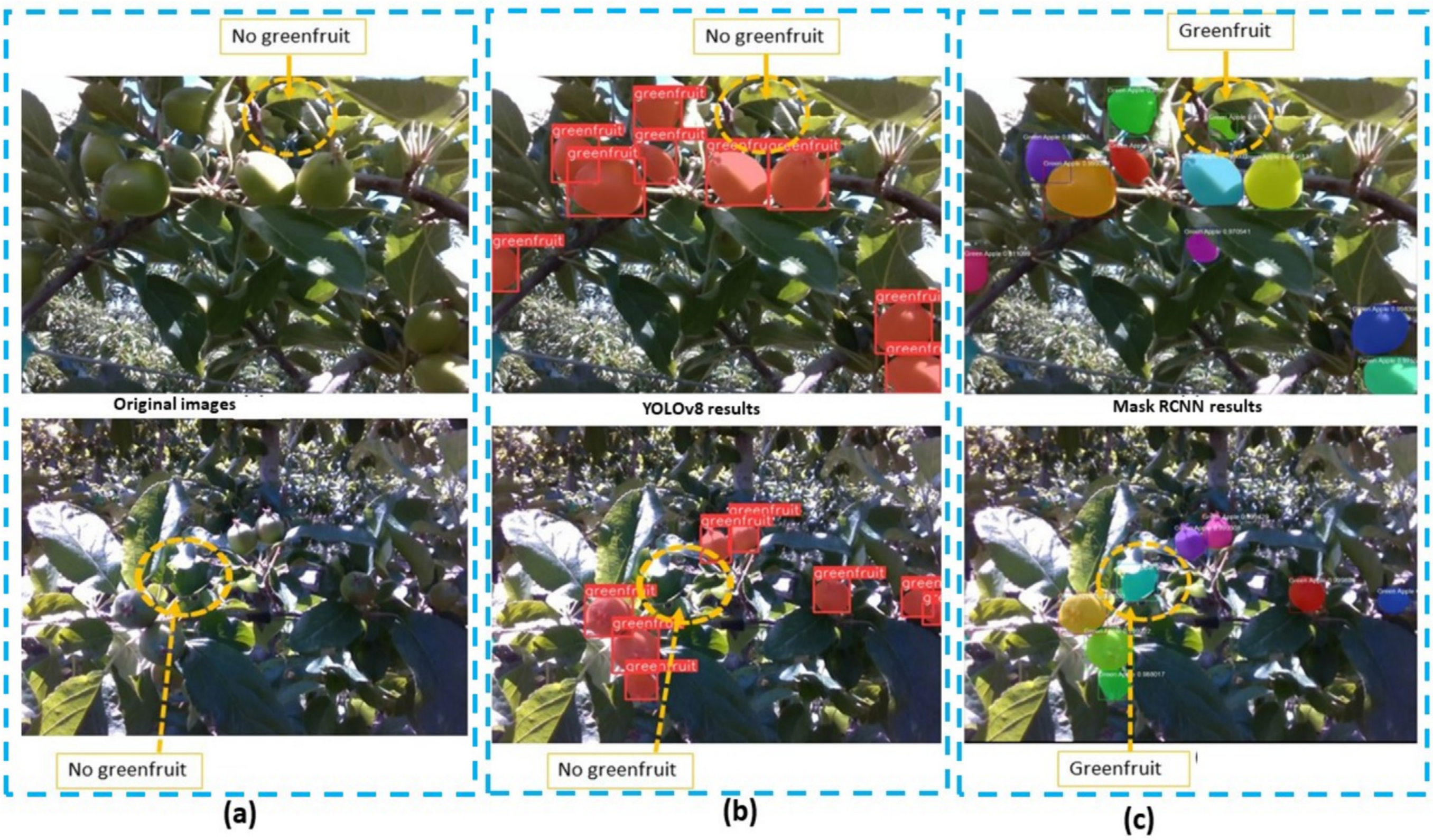

**Fig. 8.** Example images showing the performance of two methods in segmenting immature green fruit in orchard condition; (a) Original images; (b) Instance segmentation results of YOLOv8; and (c) Instance segmentation results of Mask R-CNN. It is noted that some problematic regions in the canopy images (yellow circles) were incorrectly segmented as green fruit by Mask R-CNN but were correctly left as background by YOLOv8.

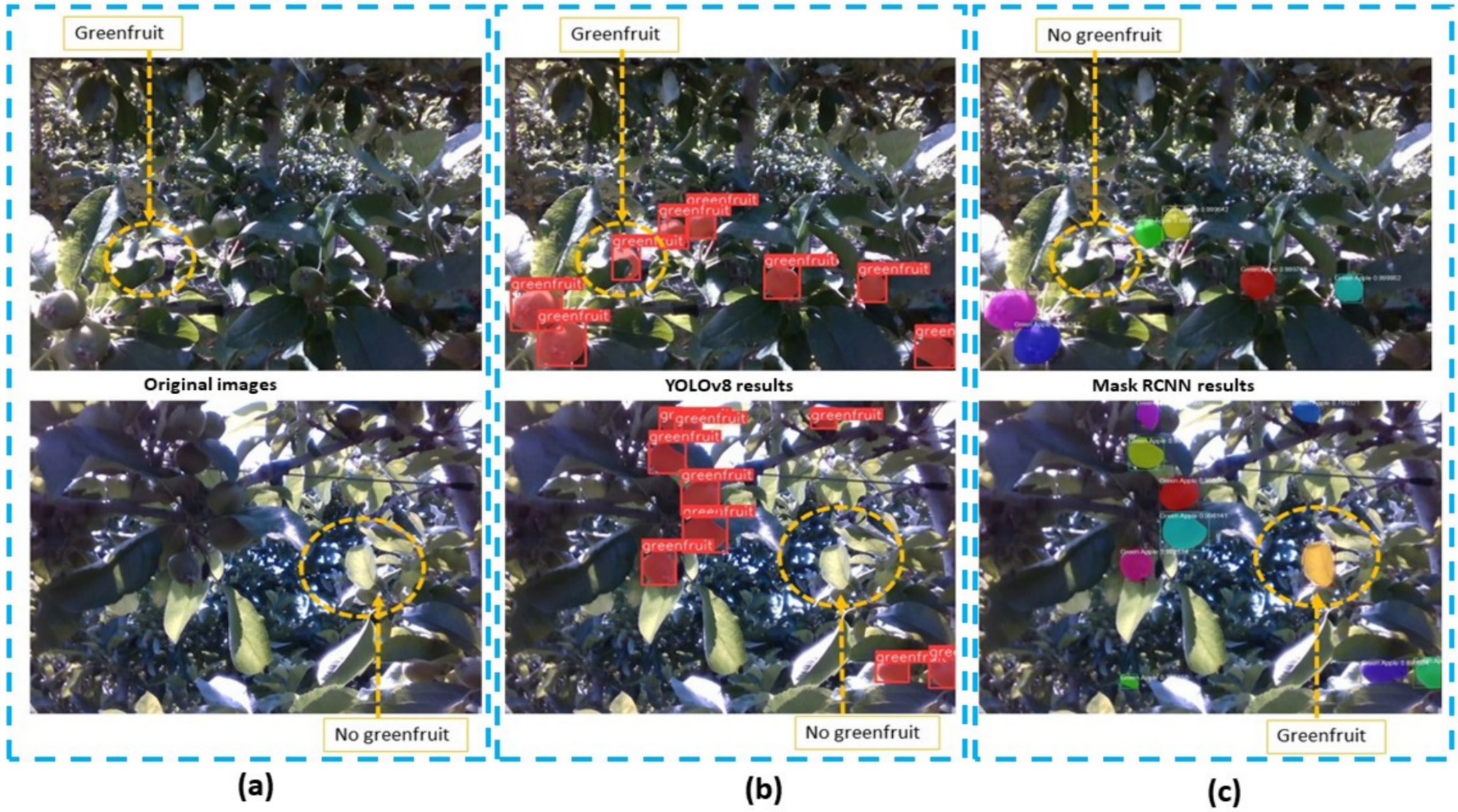

**Fig. 9.** Figure illustrating wrong detection during the growing season orchard condition, yellow region includes the focus area (a) Original image 1; Instance segmentation results of YOLOv8; and (c) Instance segmentation results of Mask R-CNN.

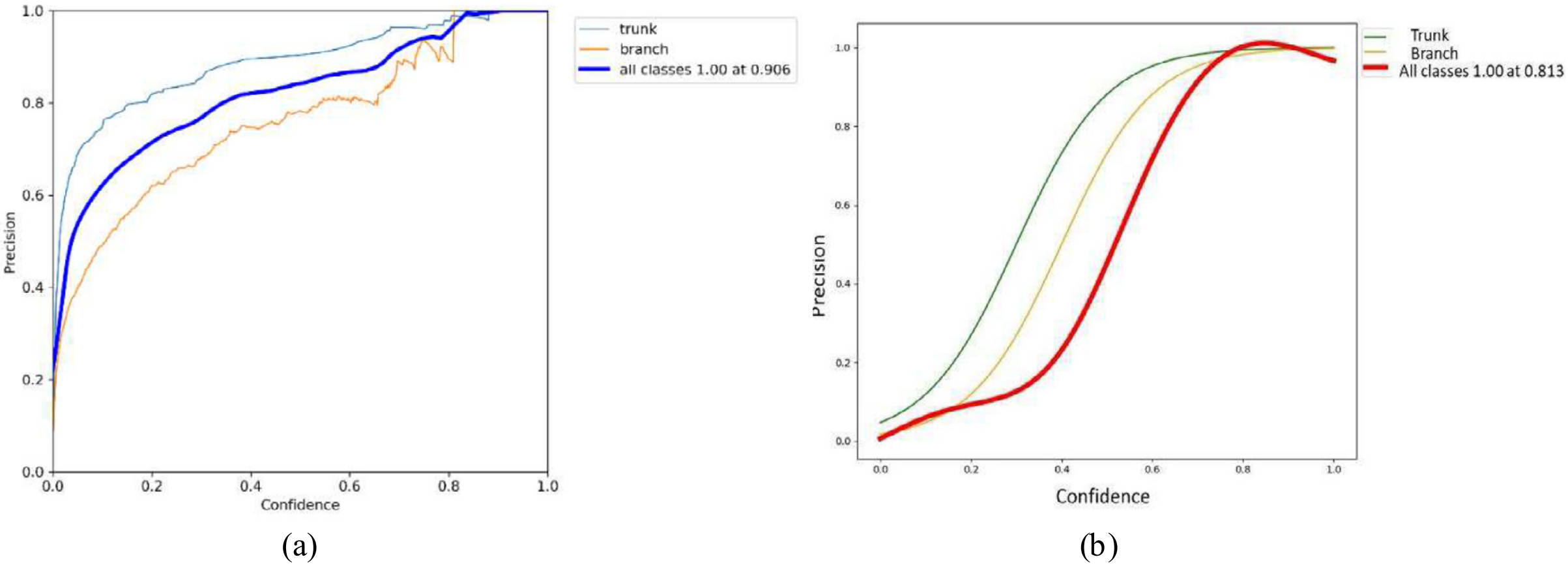

**Fig. 10.** Precision-Confidence curve for multi-class segmentation of Trunk and Branch; (a) YoloV8, (b) Mask R-CNN.

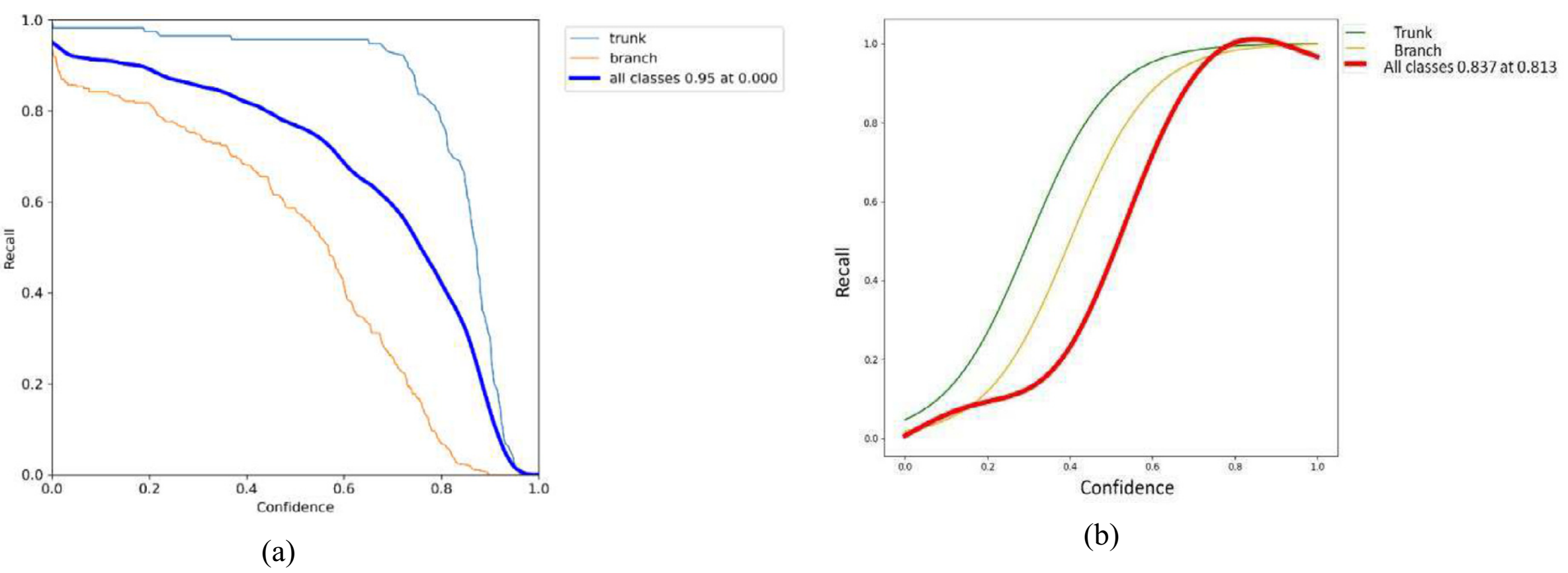

**Fig. 11.** Recall-confidence curve for multi-class segmentation of trunks and branches achieved with; (a) YOLOv8; and (b) Mask R-CNN.

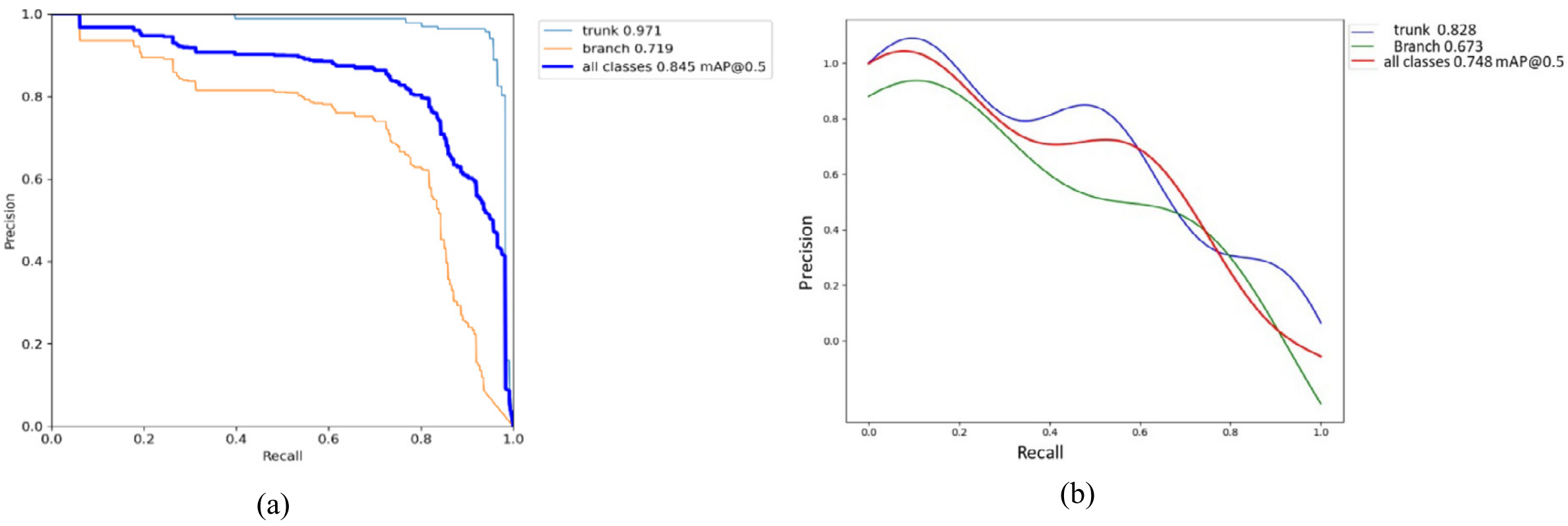

**Fig. 12.** Precision-Recall curve for multi-class segmentation of trunks and branches of dormant apple trees at mAP@0.5; (a) with YOLOv8; and (b) with Mask R-CNN.

Improved Faster-R-CNN, and achieved a mAP% of 84.89%, 82.56% and 86.41% respectively. Based on the performance measures achieved in this study (e.g., 90.2% for YOLOv8 and 85% for Mask R-CNN) for single class datasets, it was observed that YOLOv8 and Mask R-CNN has a potential to achieve a substantially better performance compared to the same achieved with the other models. However, further study to compare the performance of all these models with the same dataset would be essential to further substantiate this finding.

### *4.2. Multi-class object segmentation in images of dormant apple trees*

Similar to single class object segmentation discussed above, YOLOv8 performed better than Mask R-CNN in segmenting dormant apple tree images into multiple object classes (trunks and branches). YOLOv8 achieved a precision of 1.00 at a confidence threshold of 0.906, as shown in Fig. 10. Similarly, Fig. 11 shows that the recall for YOLOv8 reached 0.95 at the minimal confidence threshold, indicating a high degree of accuracy in segmenting these complex structures of dormant tree canopies. Mask R-CNN reached a precision of 1.00 at a lower confidence threshold of 0.813, suggesting a strong ability to correctly detecting target objects at this level of confidence (Fig. 10b). Additionally, the recall of Mask R-CNN, as depicted in Fig. 11, achieved 0.837 at the lowest confidence threshold, indicating slightly higher rate of false negatives compared to YOLOv8. Similarly, precision-recall curve (Fig. 12a) showed that YOLOv8 achieved a mean average precision (mAP) of 0.845 over all object classes at an intersection over union (IoU) of 0.5, which for the trunk and branch classes were 0.971 and 0.719, respectively. Mask R-CNN achieved relatively lower performance in multi-class segmentation tasks as well. As seen in Fig. 12b (precision-recall curve) the model achieved an all-class mAP of 0.748 at an IoU of 0.5, with individual mAP of 0.828 for trunk segmentation and 0.673 for branch segmentation.

Example images demonstrating comparative successes and failures of these models (YOLOv8 and Mask R-CNN) in segmenting trunks and branches are depicted in Figs. 13 and 14. As shown before with mAP and other measures, trunks were segmented with higher accuracy by YOLOv8 compared to Mask R-CNN, which are indicated by sample cases shown in shown in Fig. 13b and c respectively. Specifically, the branch highlighted within the yellow dotted rectangle (Fig. 13 a, b and c) was successfully detected by YOLOv8 but not by Mask R-CNN, showing YOLOv8's better performance in low light conditions compared to Mask R-CNN. The example in Fig. 13 shows that YOLOv8 was more effective in segmenting trunks. Similarly, Fig. 14 presented examples of successful and failed segmentations in both trunk and branches, which showed that YOLOv8 was more precise (less false detection) than Mask R-CNN, particularly in area with challenging lighting and complex backgrounds (e.g., a rectangular box in Fig. 14b). Comparatively, Mask R-CNN exhibited lower performance under these conditions, with the limitations being more apparent in poorly lit areas with complex backgrounds (e.g., Fig. 14c). The segmentation of the branch within the yellow rectangle (Fig. 14d) also highlighted YOLOv8's ability to detect features despite variable lighting conditions created by shadows and hue variations, an area where Mask R-CNN was less robust in segmenting desired objects (Fig. 14e).

Computational speed is one of the major performance major of these models, particularly when they are used for real-time field applications such as robotic pruning or thinning. The inference times (processing time per image during testing) required for segmenting green fruit and multi-class objects (trunks and branches) with YOLOv8 and Mask R-CNN models are presented in Table 3. It was found that YOLOv8 took only 7.8 ms to complete single-class segmentation and 10.9 ms for multi-class segmentation per test image using Intel Xeon® W-2155 CPU @ 3.30 GHz x20 processor, NVDIA TITAN Xp Collector's edition/PCIe/SSE2 graphics card, 31.1 gigabyte memory, and Ubuntu 16.04 LTS 64-bit operating system. These inference times correspond to inference speeds of approximately 128 FPS and 92 FPS, respectively for single and multi-class segmentations. Comparatively, the inference times for Mask R-CNN was higher at 12.8 ms for single class segmentation, which translates to an inference speed of approximately 78 FPS. For multi-class segmentation, the inference time increased to 15.6 ms for Mask R-CNN, or roughly 64 FPS. This difference in processing time showed suitability of the YOLOv8 for both single and multi-object instance segmentation for real-time application (summarized in Fig. 15).

According to a most recent study on comparative study of immature green apple detection using machine learning and deep learning models (Liu et al., 2022), Fully Convolutional One-Stage (FCOS) with a ResNet101 RFPN backbone achieved a precision of 81.2%. SSD, utilizing VGG16, had a precision of 69%, while YOLOv3 with Darknet-53 reached 71.3%. Faster-R-CNN and RetinaNet, both employing ResNet101-FPN, achieved precisions of 72.1% and 76.6%, respectively. Lastly, CenterNet, using the Hourglass-104 backbone, recorded a precision of 71.2%. Compared to this result, the precision recorded by both Yolov8 and Mask

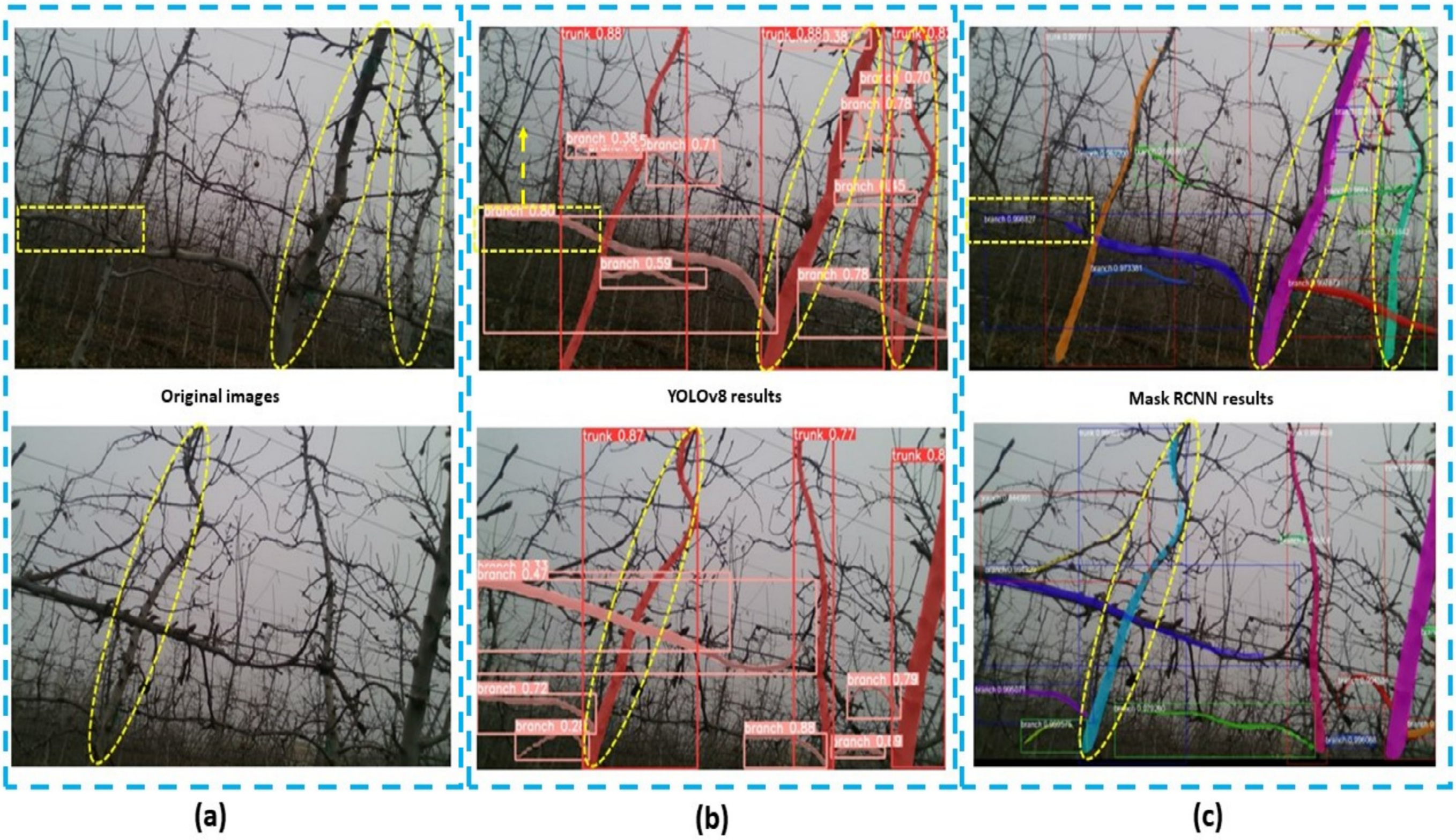

**Fig. 13.** Example results for multiclass segmentation of trunks (yellow circle) and branches (yellow rectangle) in dormant season orchard images; (a) Original images; (b) YOLOv8 segmentation results; and (c) Mask R-CNN segmentation results. This example showed slightly weaker segmentation performance of Mask R-CNN, qualitatively, compared to YOLOv8.

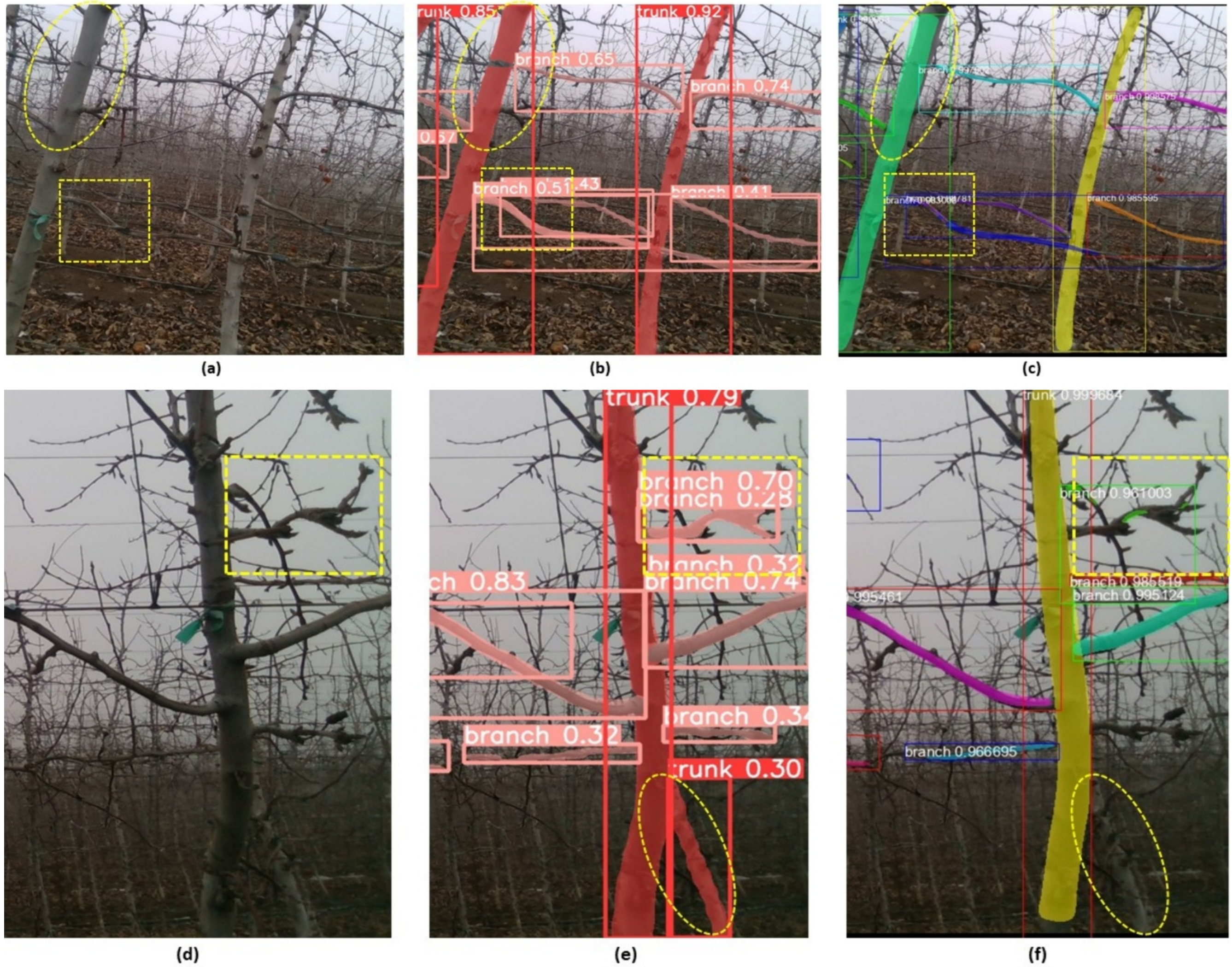

**Fig. 14.** Figures illustrating multiclass segmentation (a) Original Image1; (b) YOLOv8 segmentation (c) Mask R-CNN segmentation; (d) Original image 2; (e) Yolov8 segmentation (f) Mask R-CNN segmentation.

R-CNN (92.9 and 84.7) in this study is higher. Likewise, other recent studies conducted to detect and segment branch in apple trees such as Unet++ with an accuracy of 72%. Furthermore, studies(Kim et al., 2023) also aimed at similar segmentation tasks, yet their outcomes fall short of the results obtained in our study, where YOLOv8 and Mask R-CNN demonstrated higher precision rates of 90.6% and 81.3% respectively for branch and trunk segmentation.

### 4.3. Discussion

Mask R-CNN, while demonstrating commendable accuracy in segmenting complex agricultural images, has a slight disadvantage in terms of speed. In this study, for analyzing performances over green apple fruitlets during canopy season and tree trunks and branches during dormant season, Mask R-CNN achieved 78 FPS for single-class and 64 FPS for multi-class segmentation using the System76 workstation. Though this level of inference speed might be sufficient for most of the off-line applications where relatively higher computational capacity could be offered, it may pose challenges in real-time agricultural operations such as automated pruning and rapid decision-making for fertilization with limited computational resources. However, its detailed segmentation capability makes it highly suitable for applications where precision and detailed object delineation is essential.

YOLOv8 stands out for its speed, achieving 128 FPS (1.65)times faster than Mask R-CNN) for single-class and 92 FPS (1.43 time faster than Mask R-CNN) for multi-class segmentation with the same imaging and computational infrastructure used to test Mask R-CNN. Comparatively higher inference speed of this model is particularly advantageous for real-time agricultural tasks as discussed above. Its high precision and recall metrics further emphasize its robust performance across diverse environmental settings including variable light conditions. However, while YOLOv8 offers substantial improvements in speed and accuracy, it may sacrifice some granularity in segmentation compared to two-stage models like Mask R-CNN, which makes it slightly less applicable where minute detail is more critical than processing speed.

**Table 3**
Summary of the performance metrics of YOLOv8 and Mask R-CNN models including precision, recall, mAP@0.5, inference times, and FPS for single and multi-class object segmentation tasks in this study.

| Model | Precision | Recall | mAP@0.5 | Inference Time (ms) | Frames Per Second (FPS) |
|---|---|---|---|---|---|
| YOLOv8 (Single-class) | 92.9 | 97 | 0.902 | 7.8 | 128.21 |
| Mask R-CNN (Single-class) | 84.7 | 88 | 0.85 | 12.8 | 78.13 |
| YOLOv8 (Multi-class) | 90.6 | 95 | 0.74 | 10.9 | 91.74 |
| Mask R-CNN (Multi-class) | 81.3 | 83.7 | 0.700 | 15.6 | 64.10 |

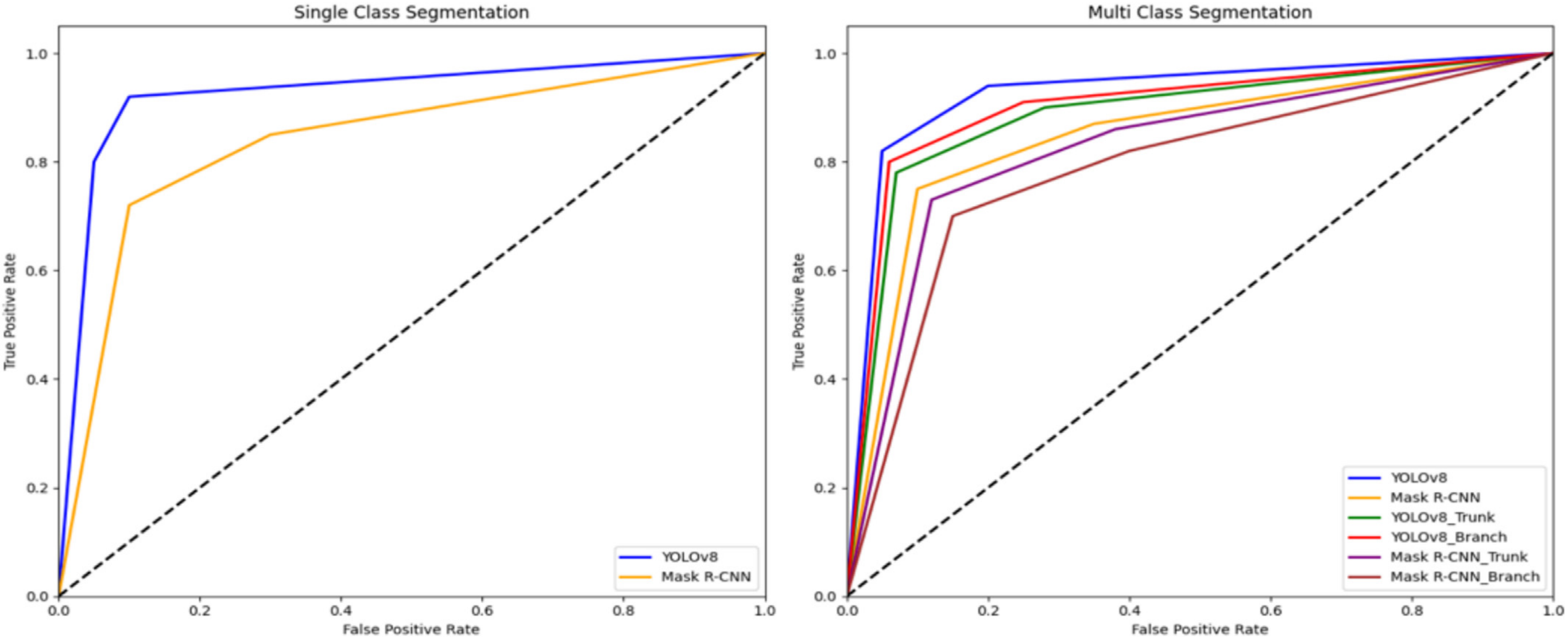

**Fig. 15.** Area under curve (AUC) for the segmentation results of both datasets: Immature green fruit (Apple) dataset on left side; Dormant season orchard dataset on right side.

YOLOv8's faster inference rates are particularly beneficial for time-sensitive tasks such as automated pruning, especially in low-light conditions, underscoring its superior suitability for operational efficiency in precision agriculture.

In general, however, these findings showed that the two models evaluated in this study could be an effective and efficient tool for developing various precision and automated agricultural tools, with potential applications extending to various crops beyond apples, which will play a crucial role in enhancing crop management and improving crop yield and quality through machine learning. Particularly, YOLOv8 showed good adaptability across different orchard conditions, which is a critical benefit in advancing robust machine learning-based solutions for future innovations in smart farming. The incorporation of machine learning is a key to meet global agricultural sustainability and food security needs.

## 5. Conclusion

In recent years, there has been increased research, development and adoption of sensing, precision, automation and robotics technologies in agricultural operations, driven by the need to minimize farming inputs including labor and increasing crop yield and quality. This study, through a comprehensive experiment in commercial orchards, provided comparative performance measures of two latest, and most widely used machine learning or deep-learning models (YOLOv8 and Mask R-CNN) for instance segmentation as it relates to their applicability to various crop monitoring and automated canopy and crop-load management tasks (e.g., automated pruning and immature green fruit thinning). Based on the results, the following specific conclusions could be made.

1. Segmentation Performance in Diverse Conditions: Both YOLOv8 and Mask R-CNN effectively segmented apple tree canopy images from both dormant and early growing seasons. YOLOv8 shows slightly better performance in environments with similar colour features between objects and backgrounds and under varying light intensities.
2. Single-Class Segmentation (Immature Green Fruit): YOLOv8 outperforms in single-class segmentation of immature green fruits, achieving a precision of 0.92 and a recall of 0.97. In comparison, Mask R-CNN exhibits slightly less effective segmentation capabilities with a precision of 0.84 and a recall of 0.88.
3. Multi-Class Segmentation (Trunk and Branch Detection): In the detection of both trunk and branches, YOLOv8 displays higher accuracy, achieving the precision and recall metrics of 0.90 and 0.95, respectively. Mask R-CNN achieved lower precision and recall, at 0.81 and 0.83 respectively, indicating reduced effectiveness in multi-class segmentation tasks.
4. Inference Speed for Multi-Class Segmentation: YOLOv8 maintains robust performance in multi-class segmentation scenarios with a speed of 91.74 FPS. In contrast, Mask R-CNN's slower inference speed of 64.10 FPS suggests limitations in handling applications requiring rapid responses.

## 6. Future work

Building on the current study, future research could focus on studying evolving capabilities of new object detection models such as YOLOv9released in February 2024 and YOLOv10 released in May 2024, and their accuracy, efficiency and adaptability to agricultural image processing. It is essential to test YOLOv9 and YOLOv10 across diverse agricultural datasets, which include various stages of crop growth, different levels of occlusion, and varying environmental conditions, to evaluate their effectiveness in actual agricultural environments. This study could particularly explore how YOLOv9 and YOLOv10 handle complex detection tasks such as identifying subtle phenotypic changes in crops under challenging light conditions or during different times of the day, situations that are typical in outdoor farming environments. Furthermore, the integration of YOLOv9 and YOLOv10 with Internet of Things (IoT) technologies could be explored to develop advanced systems for real-time monitoring and decision-making in agriculture.

## CRediT authorship contribution statement

**Ranjan Sapkota:** Writing – review & editing, Writing – original draft, Visualization, Validation, Software, Resources, Methodology,

Investigation, Formal analysis, Data curation, Conceptualization. **Dawood Ahmed:** Methodology, Formal analysis, Data curation. **Manoj Karkee:** Writing – review & editing, Supervision, Resources, Project administration, Methodology, Funding acquisition.

## Declaration of generative AI and AI-assisted technologies in the writing process

During the preparation of this work, the author used chatGPT in order to correct grammar and language. After using this tool/service, the authors reviewed and edited the content as needed and take full responsibility for the content of the publication.

## Declaration of competing interest

The authors declare that they have no known competing financial interests or personal relationships that could have appeared to influence the work reported in this paper.

## Acknowledgement

This research is funded by the National Science Foundation and United States Department of Agriculture, National Institute of Food and Agriculture through the “AI Institute for Agriculture” Program (Award No.AWD003473). The authors gratefully acknowledge Dave Allan (Allan Bros., Inc.) for providing access to the orchards during the data collection and field evaluation. Additionally, the authors would like to thank gratefully to Martin Churuvija, Sindhuja Sankaran, Christine Cromar, Bonnie Copeland and Patrick Scharf for their essential support in this project logistics.